\documentclass[letterpaper, 10pt, conference]{ieeeconf}

\IEEEoverridecommandlockouts                              

\usepackage{graphicx} 
\usepackage{amsmath} 

\usepackage{balance}

\usepackage{todonotes}

\newcommand{\citet}{\cite}

\definecolor{blueish}  {RGB}{103,135,176}
\definecolor{greenish} {RGB}{177,177,123}
\definecolor{reddish}  {RGB}{205,102,  7}
\definecolor{orangeish}{RGB}{246,160, 61}
\definecolor{blueish}  {RGB}{103,135,176}
\colorlet{darkblue}{blueish!75!black}

\usepackage{dsfont}

\newcommand{\given}{\,|\,}
\newcommand{\ang}{a}
\newcommand{\bia}{b}
\newcommand{\enc}{q}
\newcommand{\img}{z}
\newcommand{\occ}{o}

\usepackage[bookmarks=true]{hyperref}
\hypersetup{
    colorlinks,
    linkcolor={darkblue},
    citecolor={darkblue},
    urlcolor=darkblue,
}

\usepackage{hyperref}

\usepackage{cleveref}
\crefformat{footnote}{#2\footnotemark[#1]#3}

\makeatletter
\newcommand\footnoteref[1]{\protected@xdef\@thefnmark{\ref{#1}}\@footnotemark}
\makeatother

\usepackage[caption=false]{subfig}

\author{Cristina Garcia Cifuentes$^{1}$, Jan Issac$^{1}$, 
  Manuel W\"uthrich$^{1}$, Stefan Schaal$^{1,2}$ and Jeannette Bohg$^{1}$
  \thanks{$^{1}$~Autonomous Motion Department at the Max Planck Institute for 
    Intelligent Systems, T\"ubingen, Germany.
    Email: {first.lastname@tuebingen.mpg.de}}%
  \thanks{$^{2}$ Computational Learning and Motor Control lab at the
    University of Southern California, Los Angeles, CA, USA}%
}

\title{\LARGE {\bf Probabilistic Articulated Real-Time Tracking for Robot Manipulation}}

\begin{document}

\maketitle

\thispagestyle{empty}
\pagestyle{empty}


\begin{abstract}
%
We propose a probabilistic filtering method which fuses joint
measurements with depth images
to yield a precise, real-time estimate of the end-effector pose in the
camera frame.
This avoids the need for frame transformations when using it in
combination with visual object tracking methods.
 
Precision is achieved by modeling and correcting biases in the joint
measurements as well as inaccuracies in the robot model, such as poor
extrinsic camera calibration. 
We make our method computationally efficient through a principled
combination of Kalman filtering of the joint measurements and
asynchronous depth-image updates based on the \emph{Coordinate
  Particle Filter}.

We quantitatively evaluate our approach on a dataset recorded from 
a real robotic platform, annotated with ground truth from a motion
capture system.
We show that our approach is robust and accurate even under challenging
conditions such as fast motion, significant and long-term occlusions, 
and time-varying biases. 
We release the dataset along with open-source code of our
approach to allow for quantitative comparison with alternative
approaches.  
\end{abstract}



\section{Introduction}

\begin{figure}[tb]
  \vspace{-1ex}
  \centering
  \subfloat[][Apollo]{
    \includegraphics[height=87pt]{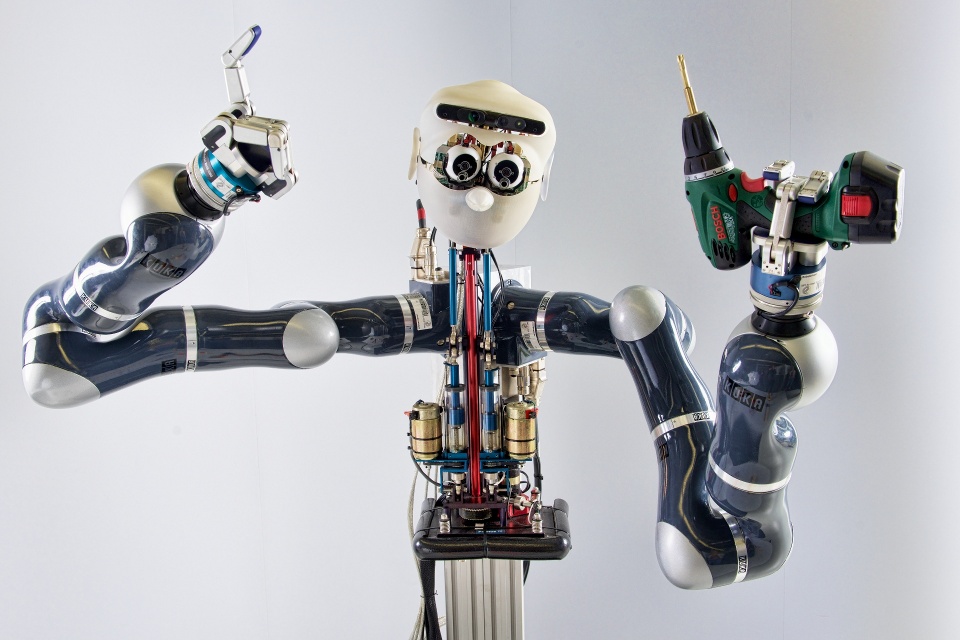}
    \label{fig:apollo}
  }
  \subfloat[][ARM]{
    \label{fig:arm}
    \includegraphics[height=87pt]{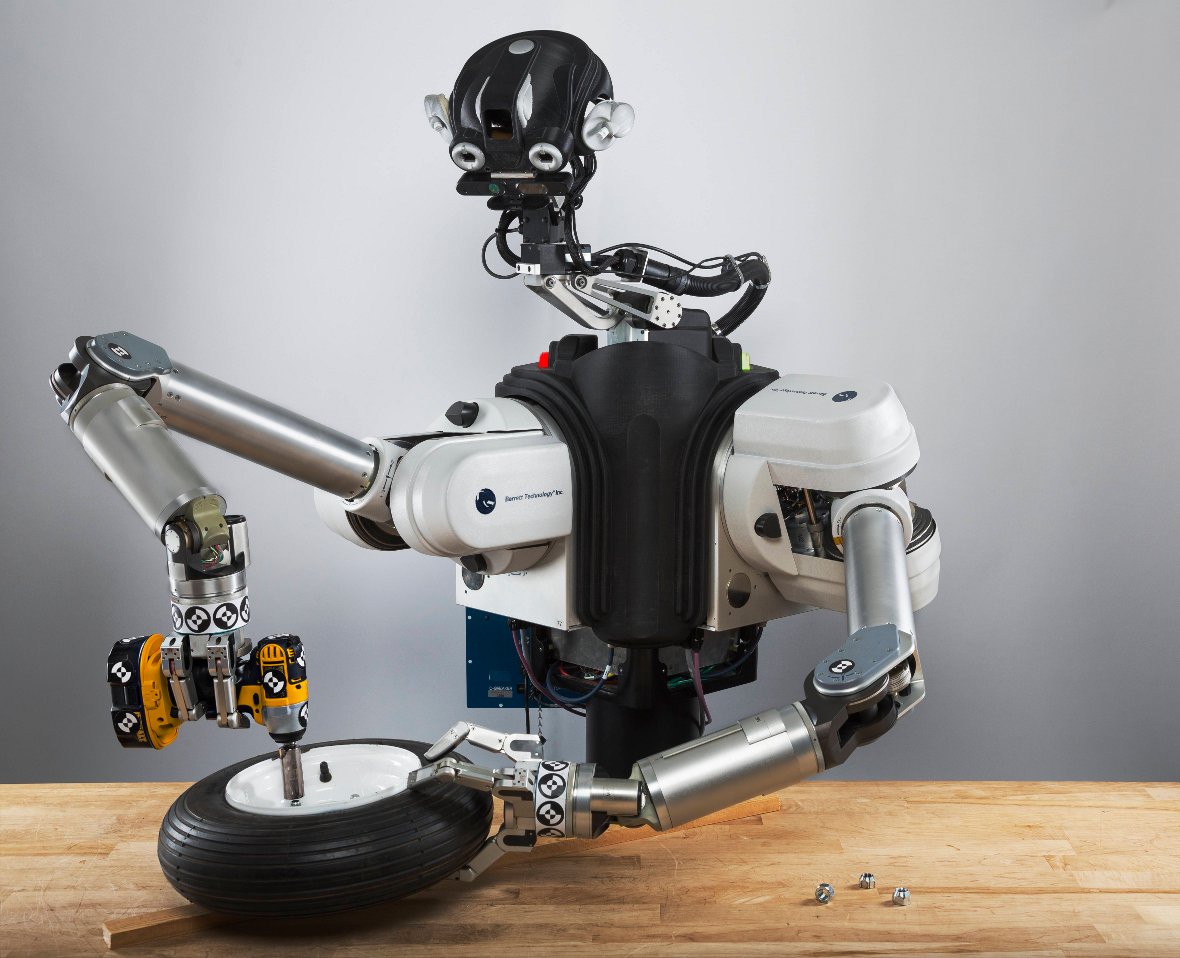}
  }\\\vspace{-1ex}
  \subfloat[][Sequence from Apollo with large simulated bias and strong occlusions]{
    \label{fig:quali-apollo}
    \includegraphics[width=0.495\linewidth]{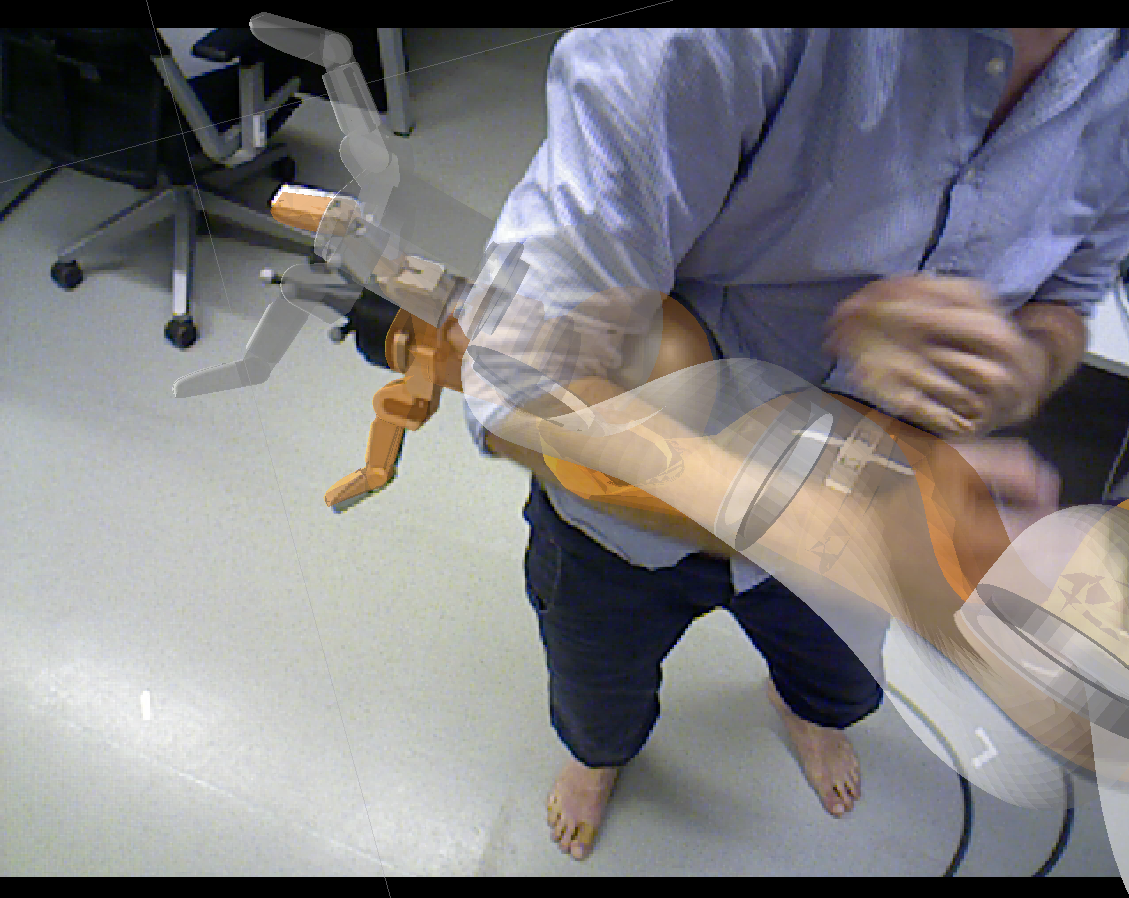}\hfill
    \includegraphics[width=0.499\linewidth]{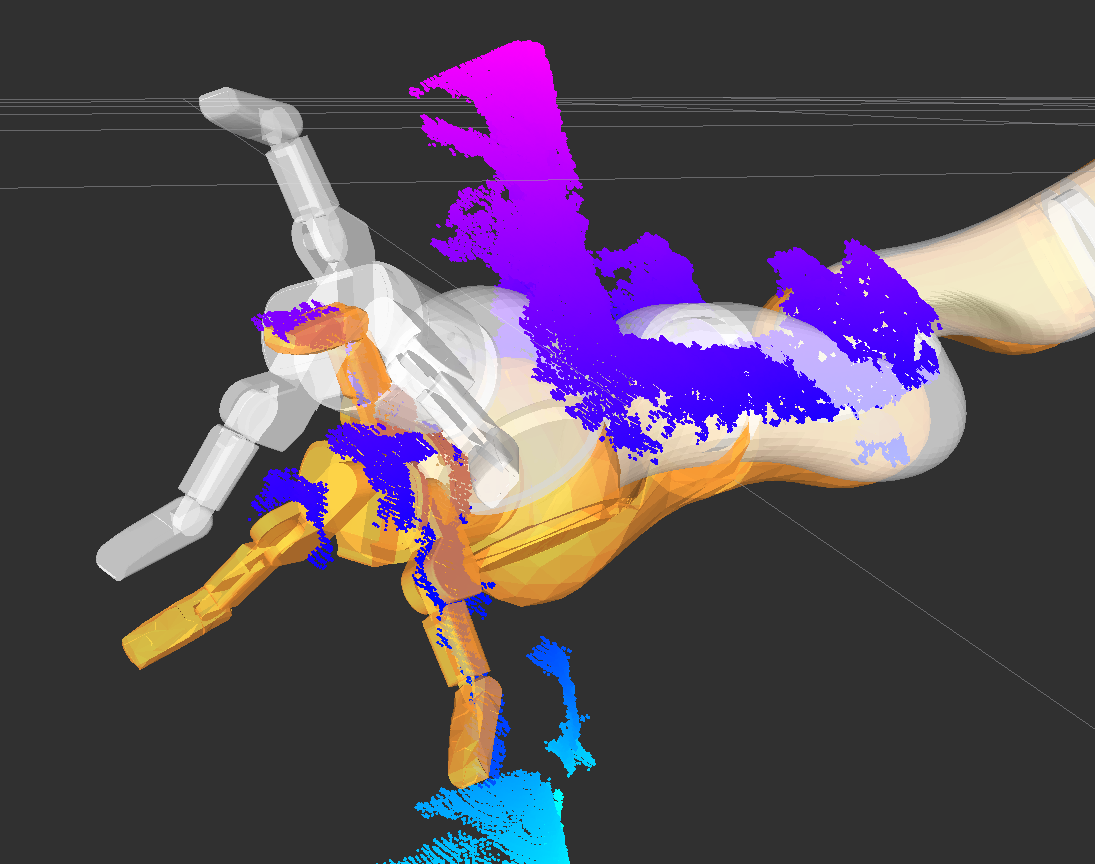}
  }\\\vspace{-1ex}
  \subfloat[][Sequence from ARM with moderate real biases]{
    \label{fiq:quali-arm}
    \includegraphics[width=\linewidth]{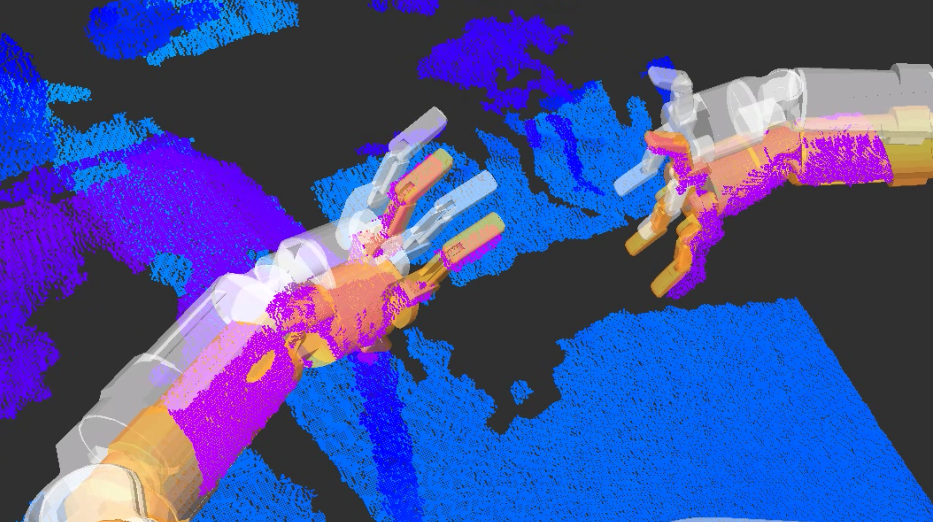}
  }
  \caption{Our method produces real-time, accurate end-effector poses in the
    camera frame, despite of joint encoder biases and/or poor extrinsic calibration
    of the camera to the robot.
    {\bf(a, b)} The two robotic platforms used
    for data recording and evaluation.
    {\bf(c, d)} We can see that for both robots our estimate (in orange) aligns  
    well with the visual information. This enables precise interaction with
    objects tracked in the same camera frame. In white: the naive forward
    kinematics, for comparison. \label{fig:quali}}
\end{figure}

Autonomous grasping and manipulation remain a frontier in robotics research,
especially in complex scenarios which are characterized by unstructured, dynamic
environments.
Under such conditions, it is impossible to accurately predict all consequences of an action far into the future.
Therefore, open-loop execution of offline-planned manipulation
actions is very likely to fail in such scenarios.

A key ingredient for accurate manipulation is
to \emph{continuously} and \emph{precisely} estimate the configuration of the
robot's manipulator and the target objects.
This configuration can be expressed as the $6$-degree-of-freedom (DoF) poses of all objects of
interest in a common frame of reference.
The focus of this paper is on the estimation of the end-effector pose with respect to the camera frame. 
We refer to this problem as \emph{robot tracking}.  

The most common approach for estimating
the end-effector pose is to apply the forward kinematics
to the joint angle measurements. 
However,
errors in the joint measurements are common, due to sensor drift or bias, and complex mechanical effects like cable stretch.
Inaccuracies in the kinematics model are quite common as well, 
because the locations of different parts of the robot with respect to each other might not 
be perfectly known.
Even slight errors in either of these two 
may lead to large errors in the end-effector pose.
On the other hand,
depth cameras can be used to accurately
estimate the end-effector pose, but images are typically received at a lower
rate, often with significant delay. Additionally, the computation required
to obtain an estimate from a depth image is typically large, which further increases
the delay until the estimate will be available.

In order to achieve precise, real-time robot tracking, 
we present a method that combines joint measurements with depth images from a camera mounted on the robot, so as to 
get the advantages from both: accurate, up-to-date estimates at a high rate.
We formulate the problem in the framework of recursive Bayesian estimation,
enabling principled, online fusion of the two sources of information. 
We account for the main sources of error along the kinematic chain by explicitly
modeling and estimating the biases of the joint measurements, and a time-varying
6-DoF transform describing a correction of the extrinsic
calibration of the camera with respect to the robot.
The algorithm we derive is computationally efficient and well-suited for
real-time implementation. We will make our code public, which can
be used off-the-shelf on any robot given its kinematic model.


For experimental validation, we collect a dataset from two humanoid robotic platforms.
It covers a range of challenging conditions, including fast arm and head motion
as well as large, long-term occlusions. We also modify the original data to
simulate large joint biases. 
We demonstrate the robustness and accuracy of our system quantitatively and
qualitatively compared to multiple baselines.
We make this dataset and evaluation code publicly available\footnotemark~to
allow evaluation and comparison of this and other methods.

\footnotetext{\url{https://git-amd.tuebingen.mpg.de/open-source/hand_tracking_dataset/wikis/home} (under construction).}

To summarize, our contributions are fourfold: 
(i) the description of a probabilistic model and a
computationally-efficient algorithm (Sections~\ref{sec:model} and~\ref{sec:algorithm});
(ii) a practically useful, real-time implementation
that we make publicly available;
(iii) experimental validation of its robustness
and accuracy (Sections~\ref{sec:setup} and~\ref{sec:results}); and
(iv) the release of a new dataset for quantitative
evaluation.

\section{Related Work}

Inaccurate and uncertain hand-eye coordination is very common for
robotics systems. 
This problem has
therefore been considered frequently in the robotics community.

One approach is to calibrate the transformation 
between hand and eye offline, prior to performing the
manipulation task. Commonly, the robot is required to hold and move
specific calibration objects in front of its
eyes~\cite{Pastor2013_calib,pauwels2016_calib,PR2_calib}. This can
however be tedious, time-consuming and the calibrated parameters may degrade over time so
that the process has to be repeated.
Instead, our approach is to continuously track the arm 
during a manipulation task, which is robust against drifting
calibration parameters or against online effects, such as cable
stretch due to increased load or contact with the environment.

One possible solution is the use of fiducial markers 
on specific parts of the robot \cite{Vahrenkamp2009}.
Markers have the advantage of being
easy to detect but the disadvantage of limiting the arm configurations
to keep the markers always in view, and requiring to precisely know their position
relative to the robot's kinematic chain. 
As an alternative to markers, different 2D visual cues can be used to track the
entire arm, at the cost of higher computational demand, e.g. 
texture, gradients or silhouettes, which are
also often used in general object
tracking~\cite{Choi_2010,Kragic_2001,Gratal_2012,HinterstoisserCISNFL12}.  
Instead, we choose as visual sensor a depth camera, which readily provides
geometric information while being less dependent on illumination effects.

In this paper, we are particularly interested in the question
of how to leverage multi-modal sensory data from e.g. proprioception as available
in a robotic system. 
%
In the following, we review
work from the class of marker-less, model-based, multi-modal tracking
methods that estimate 
the configuration of
articulated objects online and in real-time, assuming access to joint encoder readings.  

Many formulate this problem as the minimization of an objective function for each
new incoming frame, and typically use the solution from the previous time
step as initialization.

Klingensmith \emph{et al.}~\cite{Klingensmith_2013_7502} present a simple but
computationally efficient articulated {\em Iterative Closest Point\/}
(ICP)~\cite{ICP} method to estimate the joint encoder bias of
their robot arm. The objective function is defined in terms of distance between
randomly sampled 3D points on the robot model and the point cloud from
a depth camera. This is minimized by exploiting the
pseudo-inverse of the kinematic Jacobian for computing the 
gradient. 

Pauwels \emph{et al.}~\cite{pauwels_imprecise_2014} consider the problem of simultaneously estimating
the pose of multiple objects and the robot arm while it is interacting with
them; all relative to a freely moving RGB-D camera.  The objective
is defined according to the agreement between a number of visual cues (depth, motion)
when comparing observations with rendered images given the state. Instead of estimating the robot joint configuration, the authors
consider the arm as a rigid object given the encoder values, which are assumed to
be precise. To cope with remaining error in the model, robot base and
end-effector are considered single objects with the kinematics being enforced
through soft-constraints.

%
There is a related family of methods which optimize for point estimates as well,
but additionally consider a model of the temporal evolution the state, and
combine them within filtering-based algorithms.

Krainin \emph{et al.}~\cite{in_hand} propose a method for in-hand modeling of objects, which 
requires an estimate of the robot arm pose to be able to segment the
hand from the object. 
%
They perform articulated ICP similar to~\cite{Klingensmith_2013_7502}, and use
the result as a measurement of joint angles in a Kalman filter.

Hebert \emph{et al.}~\cite{hebert} estimate the pose of the
object relative to the end-effector, and the offset between the end-effector pose
according to forward kinematics and visual data. 
They consider a multitude of cues from stereo and RGB data, such as markers on
the hand, the silhouette of object and arm, and 3D point clouds. They employ
articulated ICP similar to~\cite{Klingensmith_2013_7502} or 2D signed distance
transforms similar to~\cite{Gratal_2012} for optimization. An {\em Unscented
  Kalman Filter\/} fuses the results.

Schmidt \emph{et al.}~\cite{SchmidtHNMSF15} propose a robot arm tracking method
based on an {\em Extended Kalman Filter\/}. They estimate the bias of the joint
encoders similar to~\cite{Klingensmith_2013_7502}. 
Additional to agreement
with depth images, they include physics-based constraints into their
optimization function, to penalize interpenetration between object and robot as
well as disagreement between contact sensors and the estimate.


In contrast to the filtering-based methods described above, our approach is to
model the acquisition of the actual, physical measurements, rather than 
the uncertainty associated to an optimization result.
Probability theory then
provides a well-understood framework for fusing the 
different information sources.
%
%
%
This leads to a method with few parameters which are intuitive to choose, 
because they are closely related to the physics of the sensors.


The method we propose extends our previous work on visual
tracking based on a model of raw depth images \cite{pot,cpf}.
These measurements are highly nonlinear and non-Gaussian, for which particle
filters are well suited.

In \cite{pot}, we propose a method to track the $6$-DoF pose of a rigid
object given its shape. 
Our image model explicitly takes into account occlusions due to any
other objects, which are common in real-world manipulation scenarios. 
The structure of our formulation makes it
suitable to use a Rao-Blackwellized particle filter~\cite{doucet}, which in
combination with the factorization of the image likelihood over pixels yields a
computationally efficient algorithm.

This model can be extended to articulated objects given knowledge
of the object's kinematics and shape. The additional difficulty is that
joint configurations can be quite high-dimensional
($>30$). 
In \cite{cpf}, we propose a method which alleviates this issue by sampling
dimension-wise, and show an application to robot arm tracking.

In this paper we take the latter method further, by combining the visual data
with the measurements from the joint encoders.
Due to the complementary nature of these two sensors, fusing them allows for
much more robust tracking than using only one of them.
Additionally to estimating joint angles and/or biases like ~\cite{cpf,Klingensmith_2013_7502,SchmidtHNMSF15,hebert,in_hand},
we simultaneously estimate the true camera pose relative to the
kinematic chain of the robot.
Further, differently from most approaches mentioned,
we process all joint measurements as they arrive rather than limiting ourselves
to the image rate, and we handle the delay in the images.
A crucial difference to related work is that we model not only
self-occlusion of the arm, but also occlusions due to external objects.
Although we take as input fewer cues than other methods~\cite{pauwels_imprecise_2014,SchmidtHNMSF15,hebert},
we already achieve a remarkable robustness and accuracy in robot arm tracking.
%
The dataset we propose in this paper will enable
quantitative comparison to alternative approaches. 

\begin{figure}[tb]
  \raggedleft
  \includegraphics[width=\linewidth]{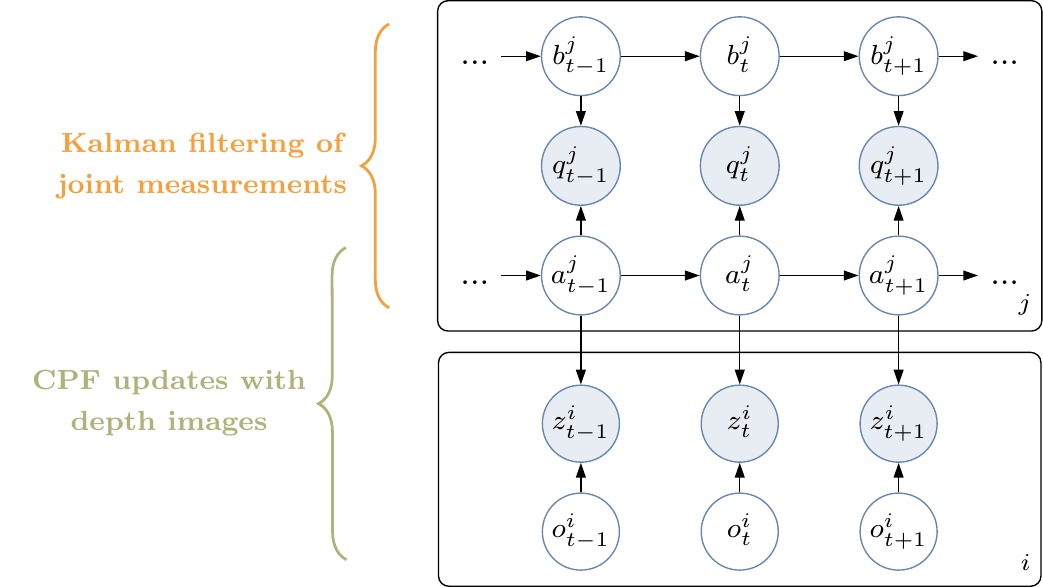}

  \vspace{3pt}
  \caption{Bayes network for our model (Section~\ref{sec:model}). Shaded nodes
    are observed (joint measurements $\enc$ and depth images $z$). White nodes
    are latent (angles $\ang$, biases $\bia$ and occlusions $\occ$). $t$, $j$ and $i$ are indices
    for discrete time, joints and pixels. Our inference algorithm (Section~\ref{sec:algorithm}) combines Kalman filtering
    of joint measurements and image updates based on the CPF \cite{cpf}. 
   \label{fig:dbrt-graph}}
\end{figure}

\begin{figure}[tb]
  \centering
  \begin{tikzpicture}
    \node (image) at (0,0) {\includegraphics[width=0.54\linewidth]{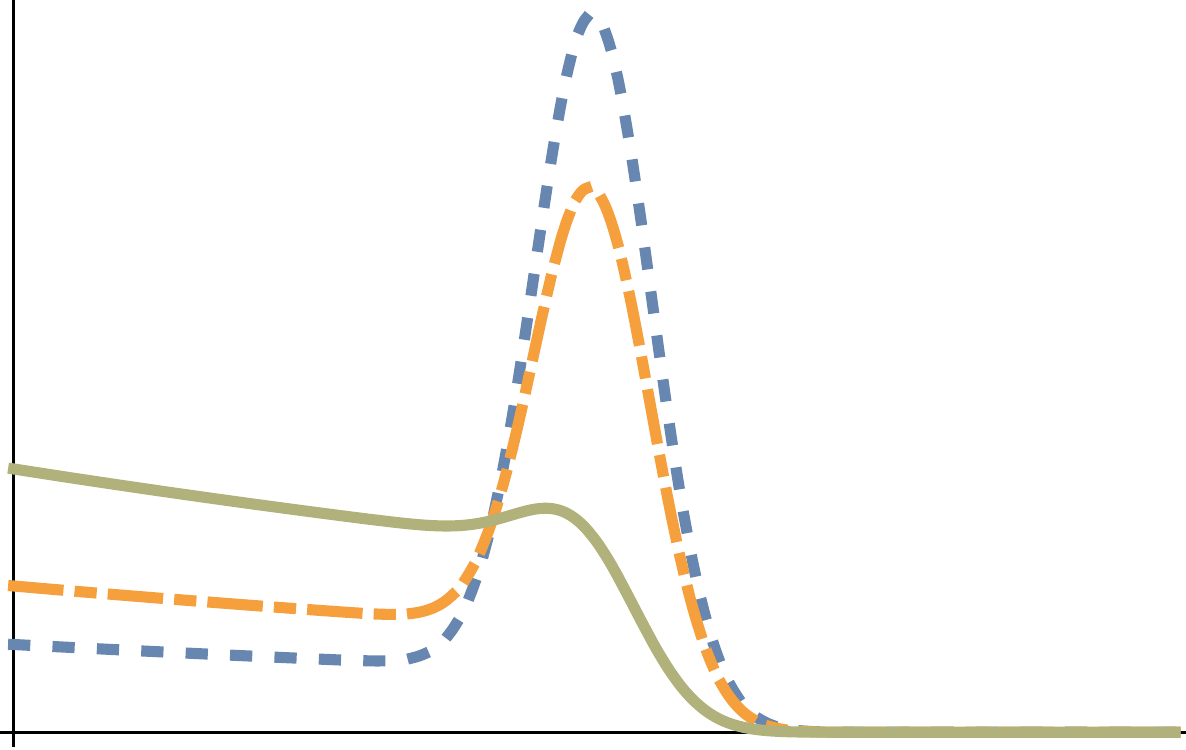}};
    \node[align = center, yshift = -1mm] at (image.south) {\small $\img_t^i$};
    \node[align = center, xshift = -2mm, rotate = 90] at (image.west) {\small $p(\img_t^i \given \ang_t)$};
   \end{tikzpicture}
  \vspace{-3mm}
  \caption{Pixel likelihood for different occlusion probabilities
    $p(o^i_t=1)$; dotted blue is lower, solid green is higher. \label{fig:img-likelihood}}
\end{figure}

\section{Modeling}
\label{sec:model}


Our goal in robot tracking is to recursively estimate 
the current joint angles $\ang_t$ of the robot
given the history of depth images $\img_t$ and joint angle
measurements $\enc_t$.
The kinematics of the robot and shape of its limbs are assumed to be
known, so the joint angles are enough to describe the full
configuration of the robot, and in particular the pose of the end
effector.
We further define a number of auxiliary latent variables, which allow
to explain the mismatch between the readings from the joint encoders
and the robot configuration observed in the depth images:
\begin{itemize}
  \item We augment the set of joints in the kinematic model with six
    extra virtual joints. These do not correspond to physical
    links, but represent a translation and a rotation between the
    nominal camera pose (i.e. as specified in the kinematic
    model\footnotemark) and the true camera pose.
    \footnotetext{This nominal camera pose can be measured in advance,
      at least roughly, if not provided by the manufacturer.}
  \item At each joint $j$, there is a bias $\bia^j_t$ that
    perturbs the joint measurement $\enc^j_t$. 
  \item For each pixel $i$ of the depth image $\img_t$, a binary
    variable $\occ^i_t$ indicates whether an external occluder is
    present.
\end{itemize}

The dependences among variables are shown in the Bayes network
in Figure~\ref{fig:dbrt-graph}.
In the remainder of this section, we define the process and measurement distributions 
that connect them, 
so as to fully specify our model. In
Section~\ref{sec:algorithm} we provide an algorithm for inference.

\subsection{Camera model}
\label{sec:depth}

In this paper, we use the same model for depth images as in \cite{pot}, which
factorizes over pixels
\begin{align}
  p(\img_t \given \ang_t, \occ_t) &= \prod\limits_i p(\img_t^i \given \ang_t, \occ_t^i). \label{eq:depth}
\end{align}
We consider three sources of error between the rendered depth $d^i(\ang_t)$ and the
measured $\img_t^i$: 
(i) inaccuracies in the mesh model, which we model as Gaussian noise; 
(ii) external occlusions, which cause the measured depth to be lower than 
the distance to the target object; and
(iii) noise in the depth sensor, which we model as a mixture of a Gaussian
distribution around the distance to the closest object and a uniform
distribution in the range of the sensor.
See \cite{pot} for the detailed expressions.

Figure~\ref{fig:img-likelihood} shows an example of the resulting pixel
likelihood after marginalizing out the occlusion variable
\begin{align}
  p(\img_t^i \given \ang_t) = \sum_{\occ_t^i = \{0,1\}} p(\img_t^i \given \ang_t, \occ_t^i) p(\occ_t^i),
\end{align}
for different occlusion probabilities $p(\occ_t^i)$. It has a peak around the rendered
distance, and a thick tail at lower distances which accounts for the possibility
of an occlusion. The peak becomes less pronounced at higher occlusion probability.

A difference between the present model (Figure~\ref{fig:dbrt-graph}) and
the model used in \cite{pot} is that here the occlusion probabilities are estimated at each
step, but not propagated over time. This is a simplification
we make to ensure tractability of the more complicated filtering problem in the present paper.
As will be shown in the experimental section, the resulting algorithm is nevertheless very
robust to occlusion.

\subsection{Joint encoder model}

We model the joint measurement as the sum of angle, bias, and independent Gaussian noise for each joint $j$:
\begin{align}
  p(\enc_t^j \given \ang_t^j, \bia_t^j) = \mathcal{N}(\enc_t^j \given  \ang_t^j + \bia_t^j, \sigma_\enc^2). \label{eq:encoder}
\end{align}

\subsection{Angle process model}

The angle of each joint $j$ follows a random walk
\begin{align}
  p(\ang_{t+1}^j \given \ang_t^j) = \mathcal{N}(\ang_{t+1}^j \given \ang_{t}^j , \Delta \sigma_\ang^2)\label{eq:angle_process}
\end{align}
where $\Delta$ is the length of the time step.\footnote{\label{foot:delta}The
  dependence on $\Delta$ arises from the integration over time of a continuous
  process with white noise. }
Parameter $\sigma_\ang$ needs to be big enough to capture fast angle dynamics.

This simple model works well in our experiments, likely because of the frequent
measurements (every 1-3~ms) with little noise. An interesting question for future 
work is whether the performance could be improved by using a more complex model,
taking into account rigid body dynamics and the control commands sent to the
robot.

\subsection{Bias process model}

We model the bias such that 
%
its variance does not grow indefinitely large in the absence of measurements.
A simple linear model that achieves this is the following random walk
\begin{align}
  p(\bia_{t+1}^j \given \bia_{t}^j) = 
  \mathcal{N}(\bia_{t+1}^j \given 
  c^\Delta \bia_{t}^j , 
  \Delta \sigma_\bia^2), \;\;\; c < 1, \label{eq:bias_process}
\end{align}
where $\sigma_b$ denotes the noise standard deviation,
$c$ is a parameter specifying how fast the process tends to $0$, and $\Delta$ is  
the time step length.\footnoteref{foot:delta}

To check that this process would behave as desired, and to gain some
intuition on how to choose the parameters, it is helpful to look at
its asymptotic behavior in the absence of measurements.
We can obtain the asymptotic distribution 
by taking the distributions at two consecutive time steps
\begin{align}
  p(\bia_t^j) &= \mathcal{N}(\bia_t^j \given \mu_*, \sigma_*^2) \\
  p(\bia_{t+1}^j) &= \mathcal{N}(\bia_{t+1}^j \given c^\Delta \mu_*, \Delta \sigma_\bia^2 + c^{2\Delta} \sigma_*^2)
\end{align}
and equating their means and variances, which yields 
\begin{align}
  \mu_* &= 0 
  ; \;\;\;
  \sigma_*^2 = \frac{\Delta}{1 - c^{2\Delta}} \sigma_\bia^2 \label{eq:bias_variance}
\end{align}
for any $\sigma_b > 0$ and $c<1$. 
%
We can see that in the absence of measurements 
the mean of the bias converges to zero (which is reasonable when there is no data suggesting otherwise),
and the variance converges to some constant which 
depends on the choice of $c$ and $\sigma_b$.
Equation~\eqref{eq:bias_variance} can help us find meaningful values
for these parameters.

\section{Algorithm}
\label{sec:algorithm}

Having defined our process and measurement models, our goal now is to find an
algorithm for recursive inference. That is, we want to obtain the current belief
%
%
%
$p(\ang_t, \bia_t \given \img_{1:t}, \enc_{1:t})$
from the latest measurements $\img_t$ and $\enc_t$,
and the previous belief $p(\ang_{t-1}, \bia_{t-1} \given \img_{1:t-1}, \enc_{1:t-1})$.

Joint and depth measurements are typically generated at different rates
in real systems, which makes it useful to be able to separately incorporate a
joint measurement or a depth measurement to our belief at any point in time.
We assume that we receive joint measurements at a high rate, so we choose the time
interval between joint measurements to be the basic time step at which we want
to produce estimates.
Further, we have to cope with a delay in the depth image, as explained in
Section~\ref{sec-delay}.

\subsection{Filtering joint measurements}
\label{sec:filter-joints}

Defining $h_t := \{\img_{1:{t-1}}, \enc_{1:{t-1}}\}$ for brevity, 
the incorporation of a joint measurement can be written as 
\begin{align}
  p(&\ang_t, \bia_t \given \enc_{t}, h_t) \propto p(\enc_t \given \ang_t, \bia_t) \cdot {} \label{eq:KF} \\
    &\int\limits_{\ang_{t-1}, \bia_{t-1}} p(\ang_t \given \ang_{t-1}) p(\bia_t \given \bia_{t-1}) 
    p(\ang_{t-1}, \bia_{t-1} \given h_t) \nonumber 
\end{align}
where we used our assumption that the angle process \eqref{eq:angle_process} 
and the bias process \eqref{eq:bias_process} are independent.
The models involved in \eqref{eq:KF} are \eqref{eq:encoder}, 
\eqref{eq:angle_process} and \eqref{eq:bias_process}. All of them are
linear Gaussian, hence \eqref{eq:KF} has a closed-form solution, which 
corresponds to one time step of a Kalman Filter (KF) \cite{kalman1960new}.
Furthermore, all these models factorize in the joints. 
Hence, if the initial belief $p(\ang_{t-1}, \bia_{t-1} \given h_t)$
factorizes too, we can filter with an independent KF for each joint,
which greatly improves efficiency.
We will see in the following how we keep the belief Gaussian and
factorize in the joints at all times, so that this is indeed the
case.

\subsection{Updating with depth images}
\label{sec:filter-depth}

Let us define $\hat{h}_t := \{\enc_t, h_t\}$ as the history of
measurements before the image update.
Each time a depth image is received, we update the belief
obtained in \eqref{eq:KF}, i.e. $p(\ang_t, \bia_t \given \hat{h}_t)$, to get the
desired posterior $p(\ang_t, \bia_t \given \img_t, \hat{h}_t)$.

The objective in this section is to write this update in such a form that
we can apply our previous work \cite{cpf}, 
so as to handle the high-dimensional state efficiently.

We begin by noting the following equalities:
\begin{align}
  p(\ang_t, \bia_t \given \img_t, \hat{h}_t) 
  &=\frac{p(\img_t \given \ang_t, \bia_t, \hat{h}_t) p(\ang_t, \bia_t \given \hat{h}_t)}
         {p(\img_t \given \hat{h}_t)}  \label{eq:posterior1} \\ 
  &=\frac{p(\img_t \given \ang_t, \hat{h}_t) p(\ang_t \given \hat{h}_t) p(\bia_t \given \ang_t, \hat{h}_t)}
         {p(\img_t \given \hat{h}_t)}  \label{eq:posterior2} \\
  &=p(\ang_t \given \img_t, \hat{h}_t) p(\bia_t \given \ang_t, \hat{h}_t), \label{eq:posterior4}
\end{align}
where from \eqref{eq:posterior1} to \eqref{eq:posterior2}
we used that our image model does not depend on the bias. According to
\eqref{eq:posterior4}, the desired posterior can be decomposed into
two terms. 
The first is the posterior in the angle only, which we will derive in
the following.
The second is easily obtained from the prior \eqref{eq:KF} by
conditioning on $\ang_t$, which for a Gaussian can be done in closed
form \cite[p.~90]{bishop}.

\subsubsection{Posterior in the angle}

We can write the posterior in the angle as
\begin{align}
  p(&\ang_t \given \img_t, \hat{h}_t) \propto 
    p(\img_t \given \ang_t) p(\ang_t \given \hat{h}_t), \label{eq:PF} 
\end{align}
where the first term 
is the image observation model \eqref{eq:depth}, and the second
is readily obtained from the Gaussian prior \eqref{eq:KF} by marginalizing out 
$\bia_t$ \cite[p.~90]{bishop}.

Since this update only involves the image $\img_t$ and the joint angles $\ang_t$, 
we can solve it in the same
manner as we did in \cite{cpf}. This involves sampling
from $p(\ang_t \given \hat{h}_t)$ dimension-wise, 
and weighting with the likelihood $p(\img_t \given \ang_t)$.
For more details, we refer the reader to \cite{cpf, pot} --
what is important here is that 
this step creates a set of particles 
$\{^{(l)}\ang_t\}_l$
distributed according to \eqref{eq:PF}. 

\subsubsection{Gaussian approximation}

After incorporating a depth image, 
we approximate the particle belief \eqref{eq:PF} 
with a Gaussian distribution factorizing in the joints
\begin{align}
 p(\ang_t \given \img_t, \hat{h}_t) \approx 
 \prod_j \mathcal{N}(\ang_t^j \given \mu^j_t, \Sigma_t^j) \label{eq:angle_posterior}
\end{align}
Moment matching is well known to be the minimum KL-divergence
solution for Gaussian approximations \cite[pp.~505-506]{bishop}.
Therefore, we have 
\begin{align}
 \mu^j_t &= \frac{1}{L}\sum_{l=1}^L {^{(l)}a_t^j}\\
 \Sigma^j_t &= \frac{1}{L}\sum_{l=1}^L ({^{(l)}a_t^j}-\mu^j_t)^2.
\end{align}


\subsubsection{Full posterior}

Since we approximated both terms in \eqref{eq:posterior4} by Gaussians which 
factorize in the joints, the full posterior is of the same form
\begin{align}
 p(\ang_t, \bia_t \given \img_t, \hat{h}_t) = 
 \prod_j \mathcal{N}(\ang^j_t, \bia^j_t \given m^j_t, M^j_t).
\end{align}
The parameters $m_t^j$ and $M_t^j$ are easily obtained using standard Gaussian
manipulations, e.g. `completing the square' in the exponent
\cite[p.~86]{bishop}.
Hence, we can continue to filter joint measurements using independent
KFs for each joint, as mentioned in Section~\ref{sec:filter-joints}.

Approximating distributions by factorized distributions is a common
practice in machine learning to ensure tractability, in particular in
variational inference. In our case, this factorization allows us to
have $n$ KFs on a $2$-dimensional state each (with complexity $O(n)$),
instead of one KF on a $2n$-dimensional state (with complexity
$O(n^3)$).

%

\subsection{Taking into account image delay}\label{sec-delay}
An additional difficulty is that images are very data heavy,
which often leads to delays in acquisition and transmission. 
This means that at time $t$ we might receive an image with time stamp
$t' < t$. 
However, all the joint angles between $t'$ and $t$ have been
processed already, and our current belief is $p(\ang_{t}, \bia_{t} \given \enc_{t}, h_t)$.

Our solution to this problem is to maintain a buffer of beliefs and joint measurements. 
When an image is received, we find the belief $p(\ang_{t'}, \bia_{t'} \given \enc_{t'},
h_{t'})$ with the appropriate time stamp, and incorporate the image $\img_{t'}$
into this belief according to Section~\ref{sec:filter-depth}. Once this
is done, we re-filter the joint measurements in the buffer which have a time
stamp $> t'$ according to Section~\ref{sec:filter-joints} to obtain the current
belief. Therefore, the filtering of joint angles has to be extremely fast,
which is possible due to the factorization in the joints.

\subsection{Efficiency and implementation}

We implemented our tracker as two filters executed in parallel at different
rates, which can be reinitialized from each other's belief at any time as
described above.

The components which make the proposed method real-time capable are:
\begin{itemize}
\item the factorization in the pixels of our depth measurement model
  \cite{pot};
\item the use of the \emph{Coordinate Particle Filter} \cite{cpf},
  which allows to filter in the high dimensional joint space of the
  robot with relatively few particles;
\item the factorization of the belief in the joints as described in
  Section~\ref{sec:filter-depth}, which allows to apply independent Kalman
  filters for each joint; and
\item our GPU implementation for computing the depth image likelihoods for
  all particles in parallel~\cite{Pfreundt14}.
\end{itemize}

Our code can be used off-the-shelf by providing the robot's
URDF\footnote{Unified Robot Description Format; \url{http://wiki.ros.org/urdf}}
representation.
Other nice features are the optional automatic injection of the
virtual joints into the URDF model, and that it is easily configurable to account
for a roughly constant offset in the image time stamp.
%

\section{Experimental setup}
\label{sec:setup}

To evaluate our approach towards robust and accurate robot tracking, we recorded
data from two different robotic platforms.
This section describes these platforms, the type of data, baselines and evaluation
measures.
We make this dataset public, together
with the robot models and evaluation code, so as to allow for comparison among
methods and reproduction of the results here.

\subsection{Robotic platforms}
\label{sec:platforms}

We recorded data on two different
robotic platforms. They are both fixed-base
dual-arm manipulation platforms equipped with two three-fingered
Barrett Hands and an RGB-D camera (Asus Xtion) mounted on an active
head. They differ in the source of error that leads to an
inaccurate hand-eye coordination.

Apollo (Figure~\ref{fig:apollo}) is equipped with two $7$-DoF Kuka LWR IV
arms with very accurate joint encoders. 
The active humanoid head on which the RGB-D camera is mounted
has a 
mechanism consisting of two four-bar linkages that
are connected at the head and generate a coupled motion.
We use an approximation to easily 
compute the $3$ rotary DoF of the neck from linear joint encoders. This 
causes 
non-linear, configuration-dependent error in the
pose of the head-mounted camera.

The ARM robot (Figure~\ref{fig:arm}) consists of two $7$-DoF Barrett WAM arms that
are actuated using a cable-driven mechanisms with motors and encoders
in the shoulder. Upon contact with the environment or during lifting a heavy
object, the resulting cable stretch is not observable through the
encoders and therefore leads to a time-varying bias on the joint
angles.  The active head consists of two stacked pan-tilt units with
simple kinematics and accurate encoders. 

While the dominant error for Apollo comes from the inaccurate kinematic model from the base to the camera,
hand-eye coordination on the ARM robot is mostly perturbed by 
biases in the joint encoder readings. With this kind of data, we cover
the most important errors that cause inaccurate hand-eye coordination 
in many robotics systems.

\subsection{Data from Apollo}
\label{sec:data}

The data recorded on Apollo consists of time-stamped RGB-D images (although our method does not 
use the RGB images), and measurements from the joint encoders.

We also measured the pose of the robot's head and
end effector using a VICON motion-capture system. 
After a calibration procedure to align the camera frame to the VICON system, 
this data provides ground truth poses of the end effector with
regards to the camera.

The dataset has seven $60$-second-long sequences recorded while Apollo was in
gravity compensation, and where its right arm and/or head is moved by a person. The sequences
include fast motion (labeled with the `++' suffix) and simultaneous arm and head
motion (`both'). One of the sequences contains heavy, long-term occlusions of
the moving arm (`occlusion'). Another contains a long period in which the arm is
out of the camera's field of view (`in/out/in'). In five additional $30$-second-long
sequences, Apollo performs sinusoidal motion (prefix `s-') of lower, upper and full arm, head
motion, and simultaneous head and arm motion.

As mentioned in Section~\ref{sec:platforms}, the kinematics of the head are
quite inaccurate, while they are precise in the arm. Therefore, we extend this
dataset by simulating bias in the joint sensors of the arm. We do this by
adding an offset to the measured joint angles, while leaving the RGB-D images
and ground-truth poses intact. 
One version of the simulated dataset has a constant offset of $8.6$ degrees in
each arm joint. Another has a time-varying offset on each joint, consisting of smooth
steps of alternating sign and $5$ degrees of amplitude.

\subsection{Data from ARM}
In the ARM robot, the joint measurements are
contaminated with real noise and bias, but we do not have ground truth
labels. We have several sequences of depth images and joint measurements of the
robot moving its two arms in sinusoidal waves at different frequencies. We use
this data for qualitative evaluation.

\begin{figure}[!htb]
  \centering 
  \includegraphics[width=\linewidth,trim={15pt 17pt 13pt 13pt},clip]{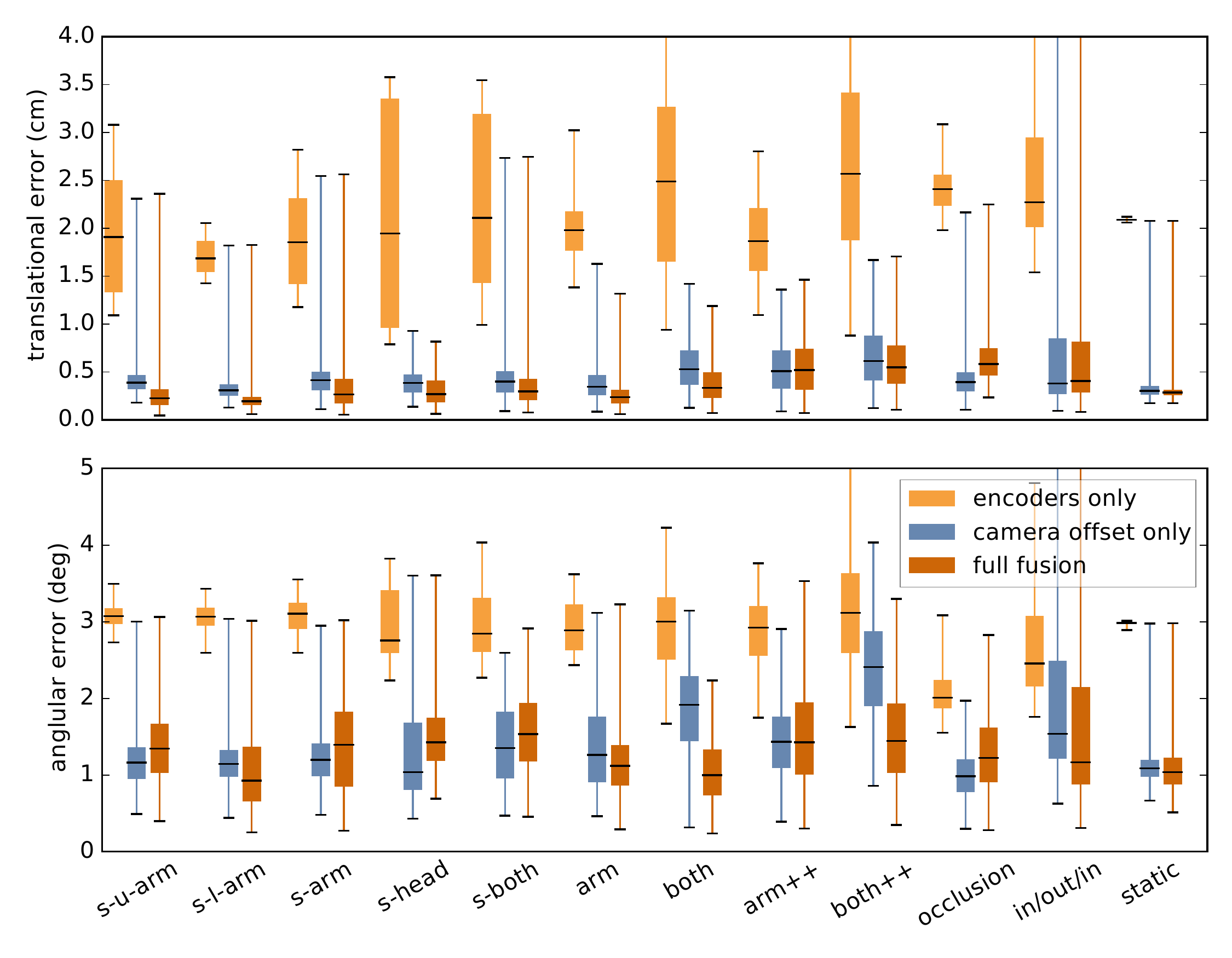}
  \vspace{-17pt}
  \caption[]{Performance on {\bf real} data with accurate joint encoders. \label{fig:box-real}}
%
\par\vspace{\intextsep}
  \includegraphics[width=\linewidth]{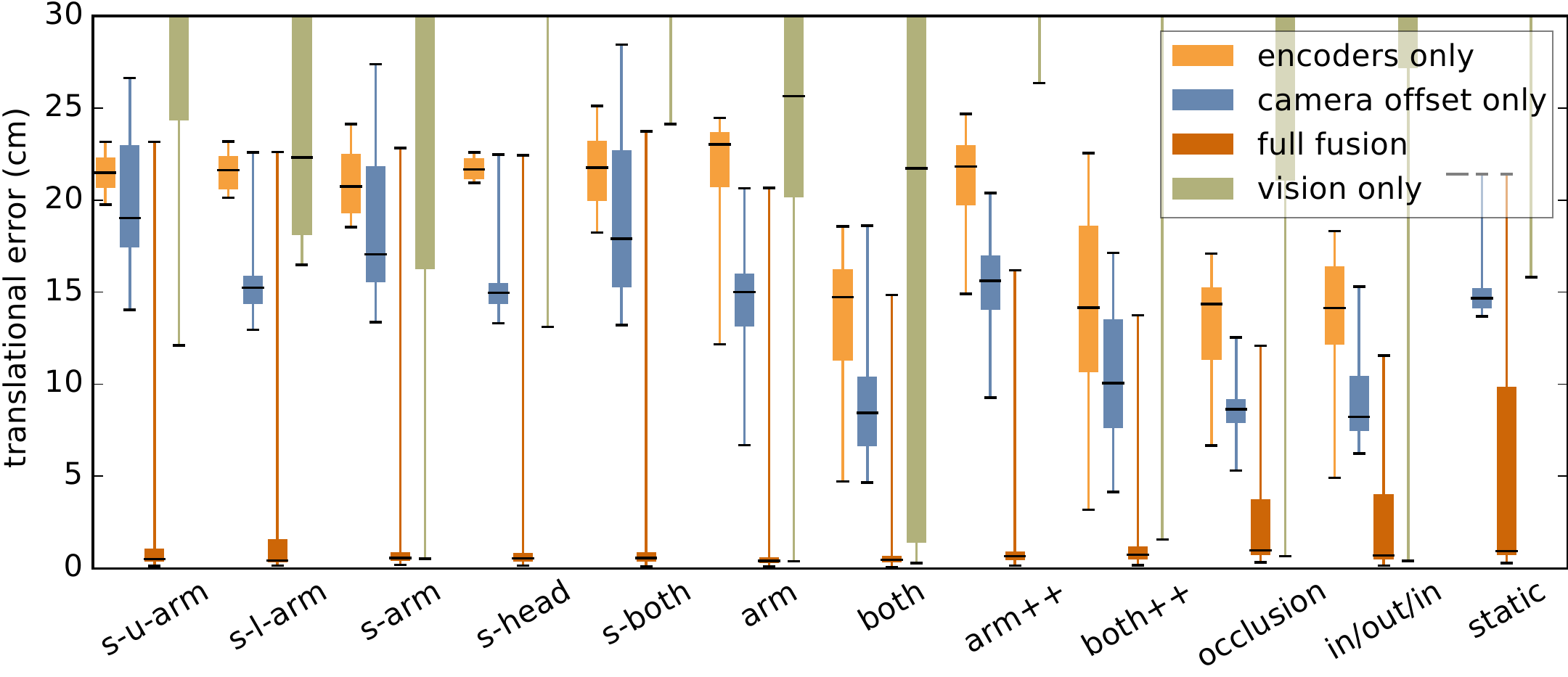}
  \vspace{-17pt}
  \caption[]{Performance on data with {\bf simulated constant}
    biases. \label{fig:box-ct}}  
%
\par\vspace{\intextsep}
%
  \includegraphics[width=\linewidth]{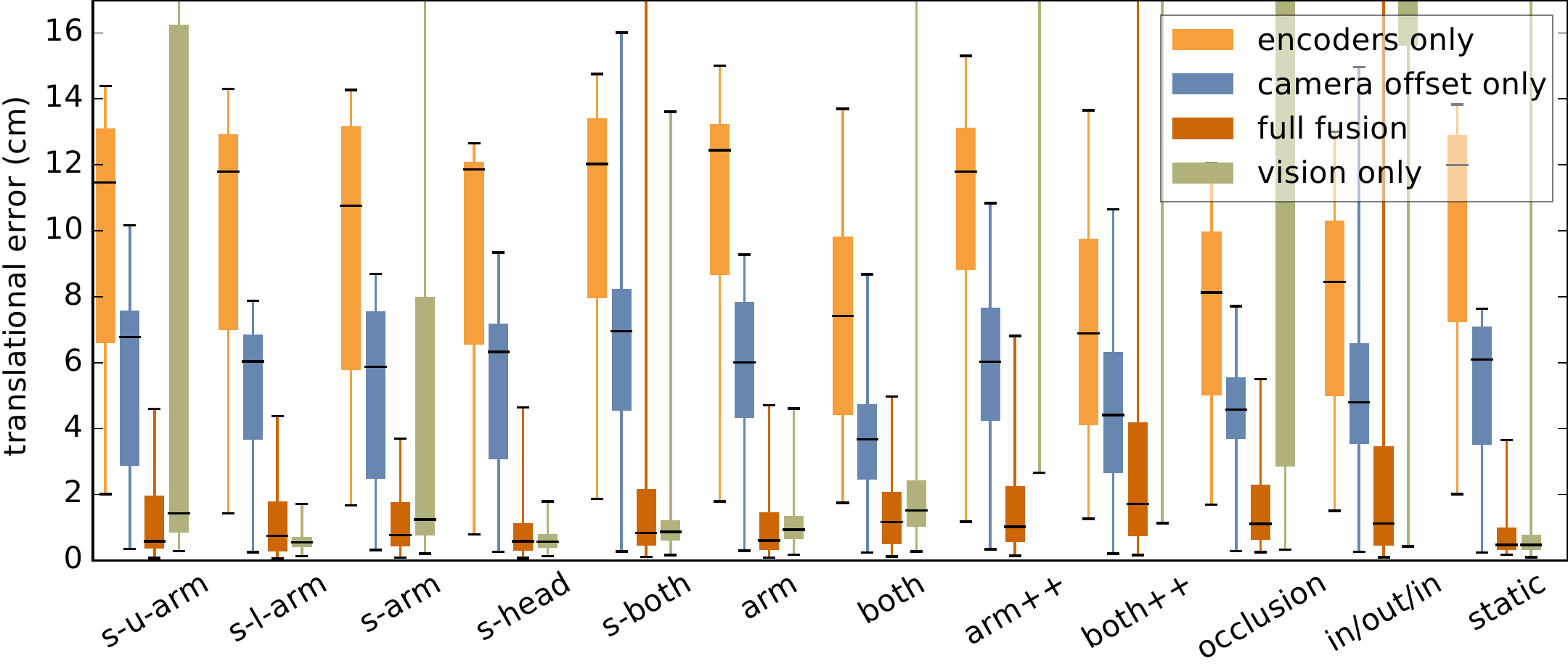}
  \vspace{-17pt}
  \caption[]{Performance on data with {\bf simulated time-varying}
    biases. \label{fig:box-var}}
\end{figure}


\begin{figure*}[!htb]
  \centering 
  \subfloat[][simultaneous head and arm motion]{
    \label{fig:time-ct-both}
    \includegraphics[width=0.325\textwidth]{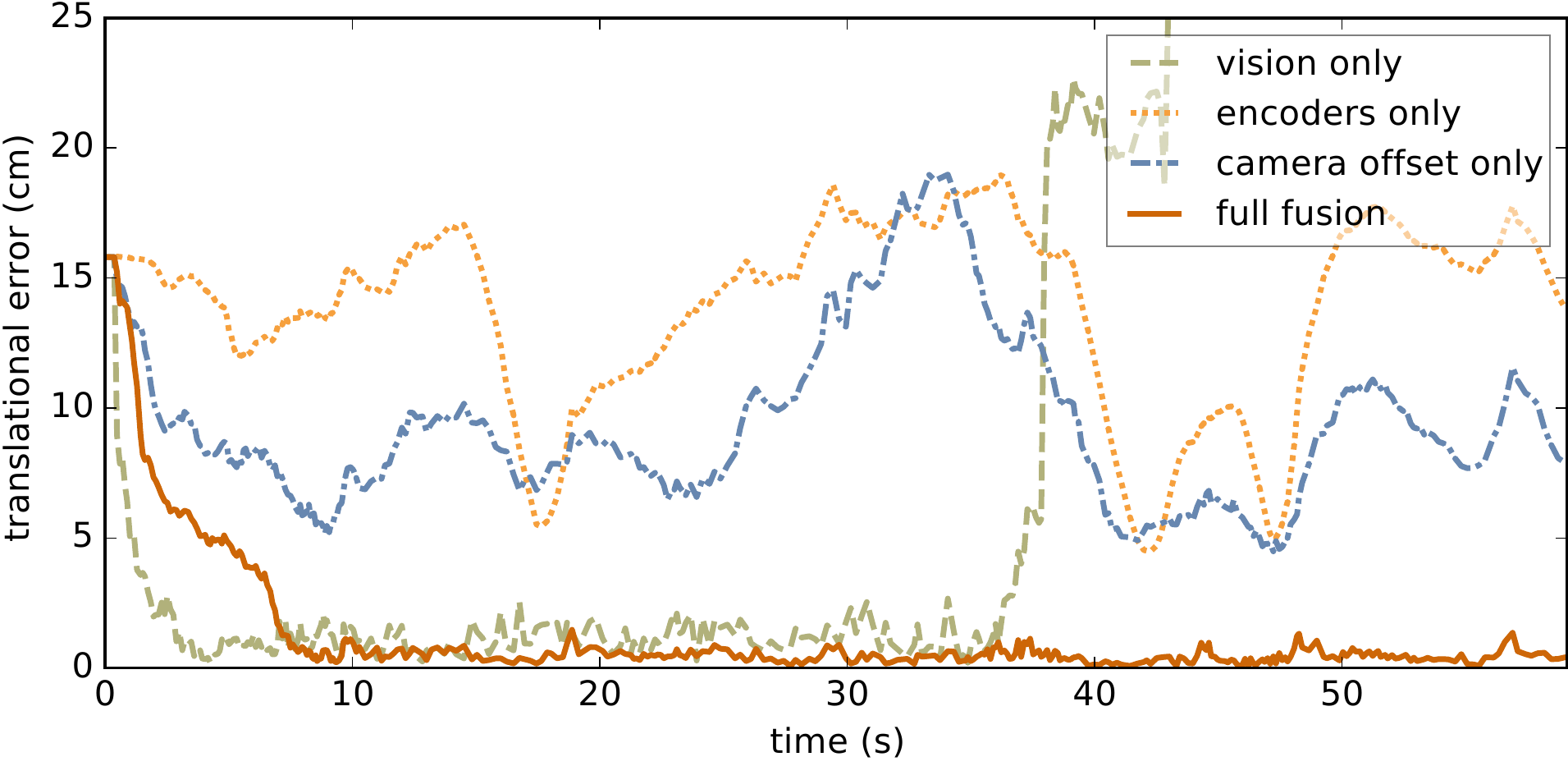}
  } \hspace{-3mm}
  \subfloat[][arm goes out of view and back]{
    \label{fig:time-ct-inout}
    \includegraphics[width=0.325\textwidth]{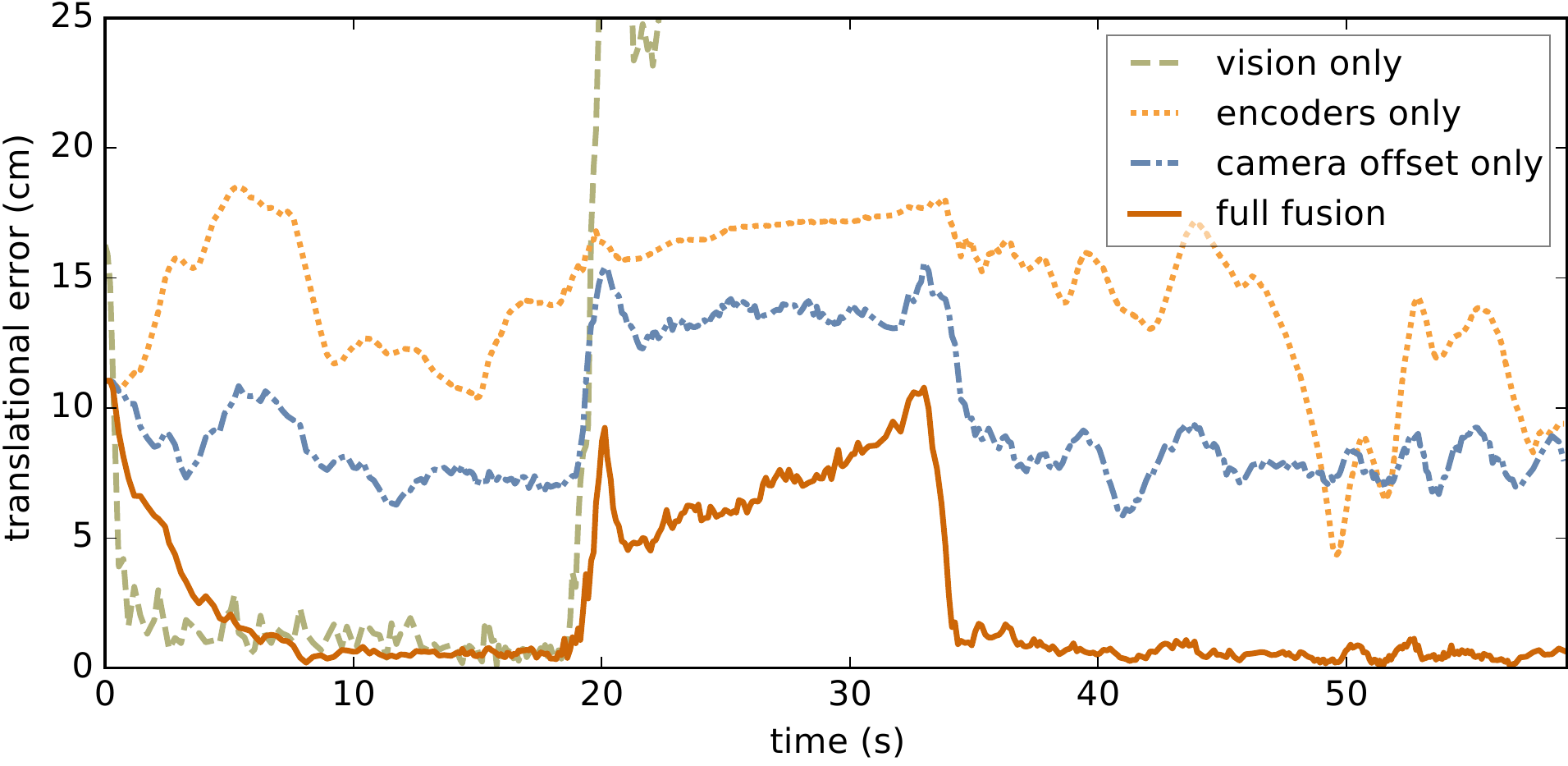}
  } \hspace{-3mm}
  \subfloat[][strong occlusions]{
    \label{fig:time-ct-occl}
    \includegraphics[width=0.325\textwidth]{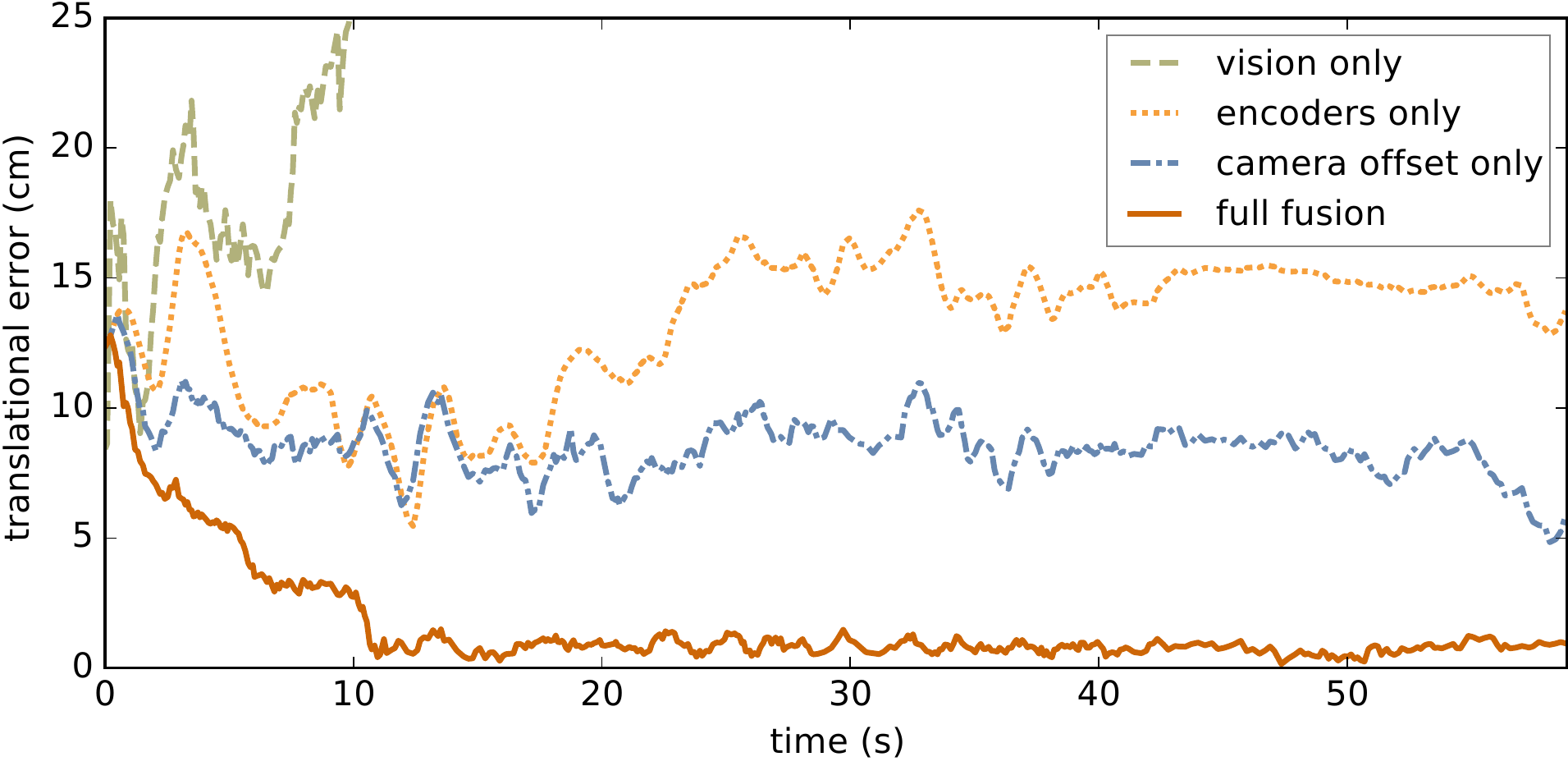}
  } \\
  \subfloat[][simultaneous head and arm motion]{
    \label{fig:time-var-both}
    \includegraphics[width=0.49\textwidth]{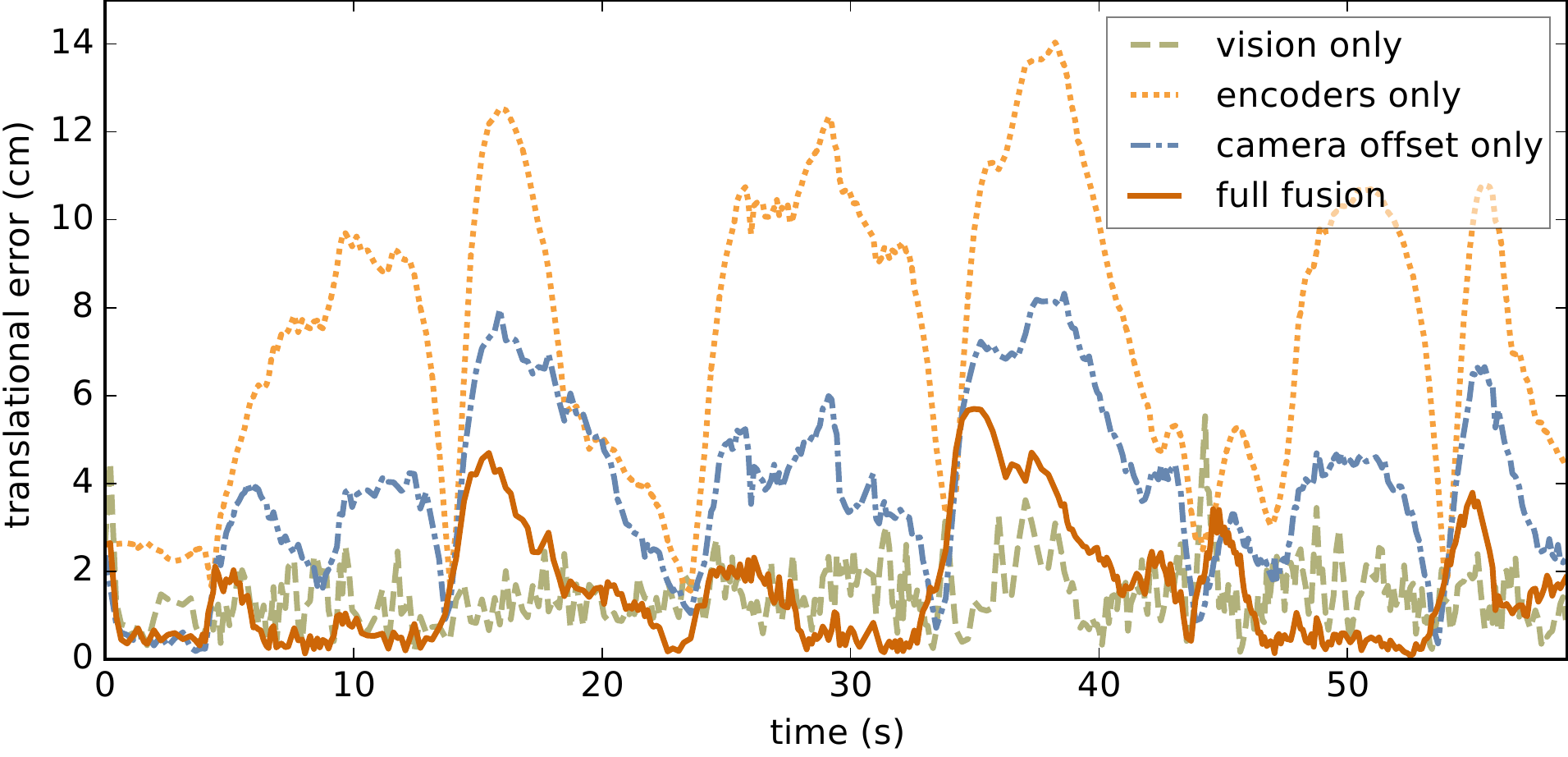}
  } \hspace{-3mm}
  \subfloat[][strong occlusions]{
    \label{fig:time-var-occl}
    \includegraphics[width=0.49\textwidth]{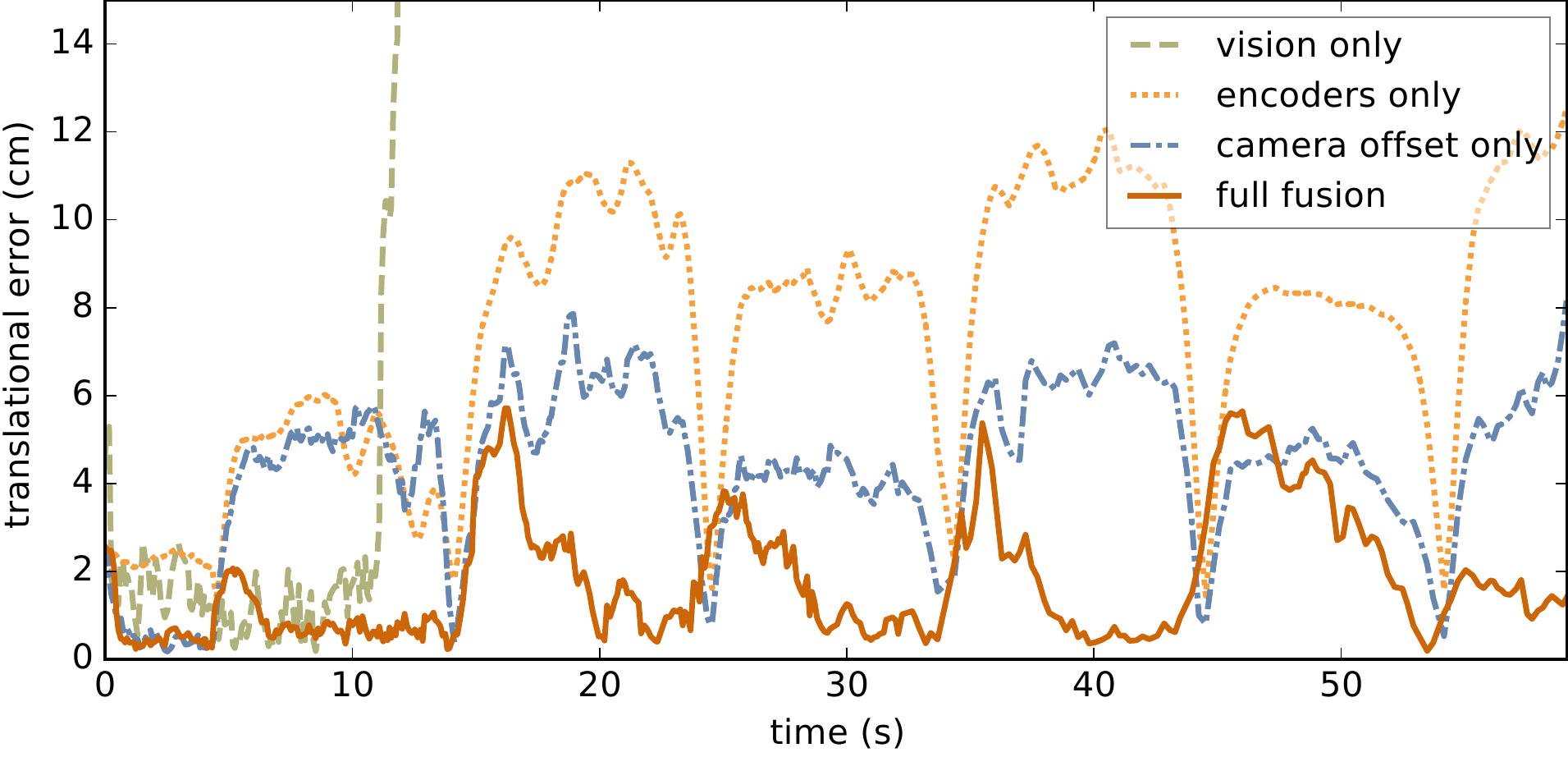}
  } 
  \caption[]{Example run on sequences with {\bf constant} (top row) and {\bf
      time-varying} (bottom row) simulated biases. \label{fig:time}}
\end{figure*}

\subsection{Performance measures}

Our method produces estimates of joint angles, and a correction of the camera
pose with regards to the robot.
In particular, this provides the pose of the estimated end effector
(the hand) in the estimated camera frame. 
We compare this pose to the ground-truth hand-to-camera pose provided by the
VICON system.
We use as performance measures the translational and angular error.

These methods will be compared in the following section:
\begin{itemize}

\item \emph{Encoders only}: a baseline that predicts the
  hand-to-camera pose by simply applying forward kinematics on the kinematic
  model, using the raw measured angles.

\item \emph{Camera offset only}: a variant of our method that assumes
  the joint measurements contain no bias, and only estimates the
  virtual joints describing the camera offset, i.e. performs online
  extrinsic calibration of the camera with regards to the kinematic
  chain. It uses both joint measurements and depth images.

\item \emph{Full fusion}: our full method, with parameters allowing to express large
  biases. It fuses joint measurements and vision at every joint, and also estimates
  virtual joints.

\item \emph{Vision only}: a variant of our method that relies only on
  images, similar to the experiments in \cite{cpf}.

\end{itemize}

To summarize the performance of each method in the dataset, we
use box plots with one bar per method and sequence. Each bar in the plot
represents statistics of the translational or angular error obtained by the
method on the sequence, taken at regular time intervals, aggregated over the
length of the sequence, and further aggregated over $10$ runs. See for example
Figure~\ref{fig:box-real}. The black ticks and limits of the colored bars
indicate the $1$st, $25$th, $50$th, $75$th and $99$th percentiles.

\subsection{Parameters}
The parameters of the image observation model are as in \cite{pot}; they 
are not specific to robot tracking. 
The new parameters in the proposed method
are the noise  $\sigma_q$ in the encoder model  \eqref{eq:encoder}, the noise
$\sigma_a$ in the angle process model \eqref{eq:angle_process}, and finally
the parameters $c$ and $\sigma_b$ of the bias process \eqref{eq:bias_process}.
All these parameters have a physical meaning and were chosen in an intuitive manner
with only minimal tuning.

There is a trade-off between the magnitude of bias the method can absorb, and
the accuracy/smoothness of the estimate. Nevertheless, we used the same
parameter set for all the quantitative experiments in the following section.  In
practice, one could achieve some improvement in accuracy by adapting the
parameters to the situation, e.g. reducing $\sigma_b$ or $c$ if we know the bias
of the encoders to be low.

\section{Evaluation}
\label{sec:results}

\subsection{Quantitative evaluation on real data}

In this experiment, we show the benefit of estimating the virtual joints to
correct inaccuracies in the head kinematics. We use real annotated data from Apollo.

Because of the properties of this robot described above, it is safe to
assume accurate joint measurements, i.e. no bias, and only estimate the virtual
joints. This correspond to the camera-offset-only method. We compare it to the
encoders-only baseline in Figure~\ref{fig:box-real}.
Clearly, estimating the camera offset decreases the error significantly.
The typical error (75th percentile) without estimating the camera offset is
in the order of a few centimeters. The camera offset correction allows
to reduce this error to a few millimeters. 
The 99th percentile is still
relatively high, after correction. This is because it takes some time 
in the beginning of a sequence for the camera offset to converge to 
the correct value. But once this is achieved, we consistently obtain 
a low error.

The full-fusion method estimates the bias in the joint angles
in addition to the camera offset. We see in Figure~\ref{fig:box-real}
that its accuracy is very similar to the camera-offset-only method.
This is expected, due to the absence of biases in the joints of the Apollo robot.

\subsection{Quantitative evaluation on simulated biases}

This section shows how our system is capable of dealing with strong,
time-varying biases on the joint measurements in a range of conditions.
We use the same data as for the previous experiment, except that here we 
corrupted the joint measurements, as described in Section~\ref{sec:data}.  

As we can see in Figure~\ref{fig:box-ct}, the error in the end-effector
position can be as high as $25$~cm when applying forward kinematics to the
biased measurements. While estimating the camera offset decreases this error by
several centimeters, we need the full fusion to bring it down to the order of
the millimeters. As shown in Figure~\ref{fig:time-ct-both}, the full-fusion method
requires some time to correct the large bias, but once it converged, it consistently
yields precise estimates.

The case of the time-varying bias (Figure~\ref{fig:box-var}) is more challenging,
because of the sudden bias changes of up to $10$ degrees. In the time-series
plots in Figure~\ref{fig:time} we can see how the fusion filter needs
some time to adapt its estimate after each change, so its error is higher
during this transition.

It is interesting to compare the behavior of the fusion tracker to the purely
visual one. The latter does not use joint measurements, so its performance does
not suffer from perturbations in them, see e.g. Figure~\ref{fig:time-var-both}.
However, it easily loses track during occlusion, fast motions, or when the robot arm is out 
of view. In contrast, the estimate of the fusion tracker 
stays at least as accurate as the encoders-only method, even when the visual information is not reliable,
e.g. during the strong occlusions in Figures~\ref{fig:time-var-occl}
and~\ref{fig:time-ct-occl}, see Figure~\ref{fig:quali-apollo} for an example frame.
Similarly, when the arm goes out of view (Figure~\ref{fig:time-ct-inout}) the
fusion tracker's estimates are pulled towards the joint measurement, so the
error increases. But then it can recover when the arm is within view
again.

\subsection{Qualitative evaluation}

Figure~\ref{fig:quali-apollo} shows one shot of the Apollo sequence with strong
occlusions and large simulated constant bias. We can see that our estimate
remains accurate despite most of the arm being covered by a person.
Figure~\ref{fiq:quali-arm} shows real depth data from the ARM robot. It gives an
idea of the amount of error produced by real bias. 

More examples of qualitative evaluation can be found in the
supplementary video (\url{https://youtu.be/YNP9UCx6Wa4}) showing robust and accurate tracking during
simultaneous arm and head motion, fast arm motion and when the arm is
going in and out of view.

\section{Conclusion}

We proposed a probabilistic filtering algorithm which fuses joint
measurements with depth images to achieve robust and accurate robot
arm tracking. 
The strength of our approach lies in modeling the
measurements in an intuitive, generative way that is realistic enough
to achieve high precision, while keeping in mind tractability. In our
case, this implied introducing and explicitly estimating auxiliary
latent variables such as pixel occlusions, encoder biases and camera
offset, which enable a good fit of the data. Some principled
approximations, as well as building on previous work such as
\cite{cpf, Pfreundt14}, made it possible to derive a computationally
efficient algorithm and real-time implementation.


In this paper, we showed that our method performs quantitatively well under the challenging
conditions of 
the dataset we propose, including fast motion, significant and long-term
occlusions, time-varying biases, and the robot arm getting in and out
of view.
Further, we have already demonstrated how this method can be
integrated into an entire manipulation system that simultaneously
tracks robot arm and object to enable pick and place tasks in
uncertain and dynamic environments~\cite{SAB_system}.

Our system is already very robust when tracking the arm using only depth
images and joint measurements. An interesting direction for future work is to
fuse data from haptic sensors, which may further improve performance when
simultaneously estimating object and arm pose during manipulation.







%
\bibliographystyle{IEEEtran}
\bibliography{bibliography.bib}

\begin{thebibliography}{10}
\providecommand{\url}[1]{#1}
\csname url@samestyle\endcsname
\providecommand{\newblock}{\relax}
\providecommand{\bibinfo}[2]{#2}
\providecommand{\BIBentrySTDinterwordspacing}{\spaceskip=0pt\relax}
\providecommand{\BIBentryALTinterwordstretchfactor}{4}
\providecommand{\BIBentryALTinterwordspacing}{\spaceskip=\fontdimen2\font plus
\BIBentryALTinterwordstretchfactor\fontdimen3\font minus
  \fontdimen4\font\relax}
\providecommand{\BIBforeignlanguage}[2]{{%
\expandafter\ifx\csname l@#1\endcsname\relax
\typeout{** WARNING: IEEEtran.bst: No hyphenation pattern has been}%
\typeout{** loaded for the language `#1'. Using the pattern for}%
\typeout{** the default language instead.}%
\else
\language=\csname l@#1\endcsname
\fi
#2}}
\providecommand{\BIBdecl}{\relax}
\BIBdecl

\bibitem{Pastor2013_calib}
P.~Pastor, M.~Kalakrishnan, J.~Binney, J.~Kelly, L.~Righetti, G.~Sukhatme, and
  S.~Schaal, ``Learning task error models for manipulation,'' in \emph{{IEEE
  Intl Conf on Robotics and Automation}}, May 2013, pp. 2612--2618.

\bibitem{pauwels2016_calib}
K.~Pauwels and D.~Kragic, ``Integrated on-line robot-camera calibration and
  object pose estimation,'' in \emph{{IEEE Intl Conf on Robotics and
  Automation}}, May 2016, pp. 2332--2339.

\bibitem{PR2_calib}
V.~Pradeep, K.~Konolige, and E.~Berger, ``Calibrating a multi-arm multi-sensor
  robot: A bundle adjustment approach,'' in \emph{International Symposium on
  Experimental Robotics (ISER)}, New Delhi, India, Dec 2010.

\bibitem{Vahrenkamp2009}
N.~Vahrenkamp, C.~B\"oge, K.~Welke, T.~Asfour, J.~Walter, and R.~Dillmann,
  ``Visual servoing for dual arm motions on a humanoid robot,'' in
  \emph{{IEEE-RAS Intl Conf on Humanoid Robots}}, 2009, pp. 208--214.

\bibitem{Choi_2010}
C.~Choi and H.~I. Christensen, ``Real-time {3D} model-based tracking using edge
  and keypoint features for robotic manipulation,'' in \emph{{IEEE Intl Conf on
  Robotics and Automation}}, May 2010, pp. 4048--4055.

\bibitem{Kragic_2001}
D.~Kragic, A.~T. Miller, and P.~K. Allen, ``Real-time tracking meets online
  grasp planning,'' in \emph{{IEEE Intl Conf on Robotics and Automation}},
  vol.~3, 2001, pp. 2460--2465.

\bibitem{Gratal_2012}
X.~Gratal, J.~Romero, J.~Bohg, and D.~Kragic, ``Visual servoing on unknown
  objects,'' \emph{Mechatronics}, vol.~22, no.~4, pp. 423 -- 435, 2012.

\bibitem{HinterstoisserCISNFL12}
S.~Hinterstoisser, C.~Cagniart, S.~Ilic, P.~F. Sturm, N.~Navab, P.~Fua, and
  V.~Lepetit, ``Gradient response maps for real-time detection of textureless
  objects,'' \emph{{IEEE} Trans. Pattern Anal. Mach. Intell.}, vol.~34, no.~5,
  pp. 876--888, 2012.

\bibitem{Klingensmith_2013_7502}
M.~Klingensmith, T.~Galluzzo, C.~Dellin, M.~Kazemi, J.~A.~D. Bagnell, and
  N.~{Pollard }, ``Closed-loop servoing using real-time markerless arm
  tracking,'' in \emph{IEEE Intl Conf on Robotics and Automation (Humanoids
  Workshop)}, May 2013.

\bibitem{ICP}
P.~Besl and N.~D. McKay, ``A method for registration of {3-D} shapes,''
  \emph{IEEE Transactions on Pattern Analysis and Machine Intelligence},
  vol.~14, no.~2, pp. 239--256, 1992.

\bibitem{pauwels_imprecise_2014}
K.~Pauwels, V.~Ivan, E.~Ros, and S.~Vijayakumar, ``Real-time object pose
  recognition and tracking with an imprecisely calibrated moving {RGB-D}
  camera,'' in \emph{{IEEE/RSJ Intl Conf on Intelligent Robots and Systems}},
  Chicago, Illinois, 2014.

\bibitem{in_hand}
M.~Krainin, P.~Henry, X.~Ren, and D.~Fox, ``{Manipulator and object tracking
  for in-hand 3D object modeling},'' \emph{{The Intl Journal of Robotics
  Research}}, 2011.

\bibitem{hebert}
P.~Hebert, N.~Hudson, J.~Ma, T.~Howard, T.~Fuchs, M.~Bajracharya, and
  J.~Burdick, ``Combined shape, appearance and silhouette for simultaneous
  manipulator and object tracking,'' in \emph{{IEEE Intl Conf on Robotics and
  Automation}}, 2012.

\bibitem{SchmidtHNMSF15}
T.~Schmidt, K.~Hertkorn, R.~A. Newcombe, Z.~Marton, M.~Suppa, and D.~Fox,
  ``Depth-based tracking with physical constraints for robot manipulation,'' in
  \emph{{IEEE Intl Conf on Robotics and Automation}}, May 2015, pp. 119--126.

\bibitem{pot}
M.~W\"uthrich, P.~Pastor, M.~Kalakrishnan, J.~Bohg, and S.~Schaal,
  ``Probabilistic object tracking using a range camera,'' in \emph{{IEEE/RSJ
  Intl Conf on Intelligent Robots and Systems}}, 2013.

\bibitem{cpf}
M.~W\"uthrich, J.~Bohg, D.~Kappler, P.~C., and S.~S., ``The {Coordinate
  Particle Filter} - a novel {Particle Filter} for high dimensional systems,''
  in \emph{{IEEE Intl Conf on Robotics and Automation}}, May 2015.

\bibitem{doucet}
A.~Doucet, N.~de~Freitas, K.~Murphy, and S.~Russell, ``{Rao-Blackwellised}
  particle filtering for dynamic bayesian networks,'' in \emph{Proc of the 16th
  Conf on Uncertainty in Artificial Intelligence}, 2000, pp. 176--183.

\bibitem{kalman1960new}
R.~E. Kalman, ``{A New Approach to Linear Filtering and Prediction Problems},''
  \emph{Transactions of the ASME - Journal of Basic Engineering}, no. 82
  (Series D), pp. 35--45, 1960.

\bibitem{bishop}
C.~M. Bishop, \emph{Pattern Recognition and Machine Learning (Information
  Science and Statistics)}.\hskip 1em plus 0.5em minus 0.4em\relax Secaucus,
  NJ, USA: Springer-Verlag New York, Inc., 2006.

\bibitem{Pfreundt14}
C.~Pfreundt, ``Probabilistic object tracking on the {GPU},'' Master's thesis,
  Karlsruhe Institute of Technology, Mar. 2014.

\bibitem{SAB_system}
J.~Bohg, D.~Kappler, F.~Meier, N.~Ratliff, J.~Mainprice, J.~Issac,
  M.~W{\"u}thrich, C.~Garcia~Cifuentes, V.~Berenz, and S.~Schaal,
  ``Interlocking perception-action loops at multiple time scales - a system
  proposal for manipulation in uncertain and dynamic environments,'' in
  \emph{Int Workshop on Robotics in the 21st Century: Challenges and Promises},
  Sep 2016.

\end{thebibliography}
%
%

\end{document}